\documentclass[sigconf, screen]{acmart}

\AtBeginDocument{%
  \providecommand\BibTeX{{%
    \normalfont B\kern-0.5em{\scshape i\kern-0.25em b}\kern-0.8em\TeX}}}

\setcopyright{acmcopyright}
\copyrightyear{2024}
\acmYear{2024}
\acmDOI{XXXXXXX.XXXXXXX}

\acmConference[XX]{Anonymous Conference 2024}{XX XX--XX,
  2024}{XX}
\acmPrice{15.00}
\acmISBN{978-1-4503-XXXX-X/18/06}

\usepackage{mycommands}

\begin{document}

\title{SignGT: Signed Attention-based Graph Transformer for Graph Representation Learning}

\author{Jinsong Chen\textsuperscript{1,2,3}, Gaichao Li$^{1,2,3}$, John E. Hopcroft$^{4}$, Kun He$^{2,3,\dag}$}

\thanks{$\dag$ Corresponding author.}
\affiliation{%
  \institution{$^{1}$ Institute of Artificial Intelligence, Huazhong University of Science and Technology
  \city{Wuhan}
  \country{China}}
  \institution{$^{2}$ School of Computer Science and Technology, Huazhong University of Science and Technology   
  \city{Wuhan}
  \country{China}}
  \institution{$^{3}$ Hopcroft Center on Computing Science, Huazhong University of Science and Technology   
  \city{Wuhan}
  \country{China}}
  \institution{$^{4}$ Department of Computer Science, Cornell University   
  \city{Ithaca}
  \country{USA}}
}
\email{{chenjinsong,gaichaolee}@hust.edu.cn, jeh@cs.cornell.edu,brooklet60@hust.edu.cn}

\def\authors{Jinsong Chen, Gaichao Li, John E. Hopcroft, Kun He}

\renewcommand{\shortauthors}{Chen, et al.}

\begin{abstract}
The emerging graph Transformers have achieved impressive performance for graph representation learning over graph neural networks (GNNs). In this work, we regard the self-attention mechanism,  the core module of graph Transformers, as a two-step aggregation operation on a fully connected graph. Due to the property of generating positive attention values, the self-attention mechanism is equal to conducting a smooth operation on all nodes, preserving the low-frequency information. However, only capturing the low-frequency information is inefficient in learning complex relations of nodes on diverse graphs, such as heterophily graphs where the high-frequency information is crucial. To this end, we propose a Signed Attention-based Graph Transformer (SignGT) to adaptively capture various frequency information from the graphs. Specifically, SignGT develops a new signed self-attention mechanism (SignSA) that produces signed attention values according to the semantic relevance of node pairs. Hence, the diverse frequency information between different node pairs could be carefully preserved. Besides, SignGT proposes a structure-aware feed-forward network (SFFN) that introduces the neighborhood bias to preserve the local topology information. In this way, SignGT could learn informative node representations from both long-range dependencies and local topology information. Extensive empirical results on both node-level and graph-level tasks indicate the superiority of SignGT against state-of-the-art graph Transformers as well as advanced GNNs.
\end{abstract}

\ccsdesc[500]{Computing methodologies~Supervised learning by classification}
\ccsdesc[300]{Mathematics of computing~Graph algorithms}

\keywords{Graph Transformer, Graph Neural Network, Graph Representation Learning, Signed Self-attention, Various Frequency Information}
%


\maketitle

\section{Introduction}
The graph is a crucial data structure that characterizes relations between entities abstracted from various domains, such as social connections in social platforms or chemical bonds in molecular graphs. 
Due to the wide application of graphs, graph representation learning, which aims to learn the informative representation of nodes on graphs to handle downstream machine learning tasks, has attracted enormous attention in recent years.
As a proverbial technique of graph representation learning, graph neural networks (GNNs) have achieved remarkable performances on diverse graph-based machine learning tasks \cite{gcn,gcnii,gin}.

Built on the message-passing mechanism~\cite{mpnn} that recursively aggregates information of immediate neighbors,  GNNs can jointly encode the graph structural features and node attribute features, demonstrating their powerful ability to learn the node representations.
Despite the prevalence, recent studies~\cite{oversmoothing,oversq,gcnii} have revealed that GNNs suffer from the inherited limitations of the message-passing mechanism, such as over-smoothing \cite{oversmoothing} and over-squashing \cite{oversq}. These limitations restrict them to capture long-range dependencies on graphs~\cite{graphtrans}, weakening their potential in graph representation learning.

On the other hand, Transformer~\cite{transformer}, as an advanced deep learning architecture, has exhibited 
powerful modeling capability in nature language processing~\cite{transformer} and computer vision~\cite{vit}. 
As the self-attention mechanism of Transformer can accurately capture the long-range dependencies of input objects, researchers have attempted to generalize Transformer for graph representation learning and develop a series of Transformer-based models~\cite{san,sat,ansgt,nodeformer,nagphormer}, called graph Transformers. 
The key idea of graph Transformers is to encode the graph topology features into the Transformer architecture and enable it 
to preserve graph structural information.
Classic strategies, such as eigenvectors of the Laplacian matrix~\cite{san,gt} and shortest-path of node pairs~\cite{graphormer}, have been adopted for strengthening the input features or the attention matrix of the Transformer.
Other recent studies~\cite{graphtrans,sat} also develop a hybrid model, which first utilizes GNN layers to capture the graph structural information and then leverages Transformer layers to preserve the long-range dependencies.
By introducing various graph information into the Transformer architecture, graph Transformers have achieved impressive performance for graph representation learning.

The self-attention module, which is the core module of graph Transformers, could be regarded as a two-step aggregation operation on the fully connected graph.
For each node, the self-attention module first calculates the attention weights of all nodes, and then utilizes the attention weights to aggregate the information of nodes. 
Due to the property of the softmax function in the self-attention module, all attention weights are positive, which could be regarded as a smoothing operation, preserving the low-frequency information of all node pairs~\cite{fagcn}.
From the perspective of spectral space, a recent work~\cite{feta} also obtains a similar conclusion that Transformer is only effective in learning the class of filters containing the low-pass characteristic.
However, existing studies \cite{fagcn,ggcn,gprgnn} show that only utilizing low-frequency information is inefficient in learning complex relations of nodes on diverse graphs.
For instance, high-frequency information is essential in learning informative node representations of heterophily graphs \cite{fagcn}, where most connected nodes belong to different labels. 
Unfortunately, the paradigm of the self-attention module in existing graph Transformers cannot adaptively preserve different frequency information for different types of graphs.

There are several modified attention-based GNNs \cite{fagcn,sign1,sign2} that develop new attention mechanisms to preserve various frequency information.  
These methods share a similar idea that utilizes the hyperbolic tangent function to replace the softmax function, omitting the normalization operation of attention values.
This strategy is not suitable for graph Transformers because the Transformer requires calculating the attention values of all nodes, which is much greater than those of GNNs only considering immediate neighbors.
If we utilize attention values without normalization to aggregate information of all nodes, a large quantity of these values could cause the vanishing gradient problem or gradient exploding problem, undermining the model training of graph Transformers.
Accordingly, a natural question arises: 
\textit{How to design a new self-attention mechanism for graph Transformer so as to capture 
various frequency information on graphs?}

\begin{figure}[t]
\centering
\includegraphics[width=8.5cm]{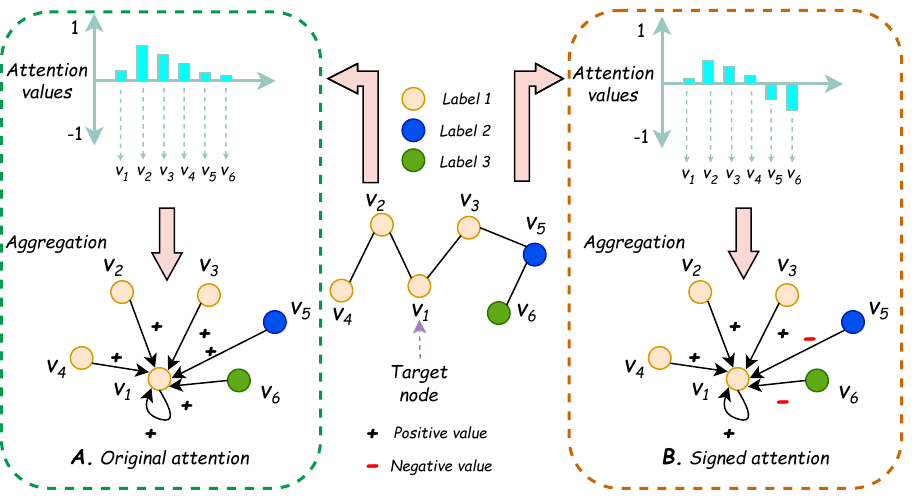}
\caption{
A toy example for illustrating the difference between the original self-attention module and our proposed signed self-attention module. 
For a given target node, the original self-attention module utilizes all positive attention values to aggregate information of other nodes.
While our proposed signed self-attention module leverages signed attention values to adaptively aggregate various frequency information of nodes from the same or different labels. 
}
\label{fig:mv}
\vspace{-1em}
\end{figure}

To answer this question, we design a new self-attention mechanism, called signed self-attention (\saname), whose key idea is to separate the sign information (which could be regarded as semantic relevance) from the attention weights in the original self-attention.
Ideally, a node pair with low semantic relevance will exhibit negative sign information, and high semantic relevance will lead to positive sign information.
In this way, \saname could adaptively preserve the low-frequency or high-frequency information of each node pair according to semantic relevance while preserving the normalization operation of attention values.
\autoref{fig:mv} illustrates the difference between our proposed \saname and the original self-attention mechanism. 

Based on \saname, we further propose a new graph Transformer, namely Signed Attention-based Graph Transformer (\name), for graph representation learning. 
First, \name leverages \saname to capture various frequency information on graphs, which could be regarded as global information. 
Then, \name develops a structure-aware feed-forward network (\sffn) by introducing the topological bias of the local neighborhood into the original feed-forward network to encode the topology features of graphs, which could be regarded as the local information.
The main advantage of \sffn is that, compared to existing graph Transformers which mainly focuses on immediate neighbors~\cite{nodeformer,gps}, \sffn can  flexibly preserve the topological information beyond immediate neighbors via the non-parametric propagation process.
By combining \saname and \sffn modules, \name can simultaneously preserve 
global long-range dependencies, as well as local topology information on graphs. 

The main contributions of this paper are summarized as follows:
\begin{itemize}

    \item We propose \saname, a novel signed self-attention mechanism that generates signed attention values based on the semantic relevance of each node pair to adaptively preserve 
    various frequency
    information on graphs.

    \item We further propose \name, a signed attention-based graph Transformer, for graph representation learning. Specifically, \name utilizes \saname to capture 
    various frequency information of the node pairs, and develops \sffn to leverage the structural bias of the neighborhood to preserve topological information.

    \item 
    Extensive experiments on both node-level and graph-level tasks show 
    the advanced performance of \name for graph representation learning, 
    which consistently demonstrates the superiority of \name against state-of-the-art graph Transformers and GNNs. 
    
\end{itemize}

\section{Related Work}
In this section, we briefly review recent works on GNNs and graph Transformers. 

\subsection{Graph Neural Network}
GNNs are famous 
models built on the massage-passing framework~\cite{mpnn} for graph representation learning.
Typical GNNs utilize the Laplacian smoothing operation~\cite{gcn,spgcn,gcnii} or the attention mechanism~\cite{gat,amgcn,spgat} to aggregate information of immediate neighbors.
To exploit the information of high-order neighborhoods, many effective strategies have been investigated.
Some methods utilize the vector concatenation operation~\cite{mixhop} or the skip connection technique~\cite{jknet} to fuse the information of high-order neighborhoods.
Other methods~\cite{sgc,appnp,gdc} leverage various propagation strategies, such as random walk~\cite{sgc} and personalized PageRank~\cite{appnp}, to capture the structural information of multi-hop neighborhoods.
Recently, several studies~\cite{fagcn,gprgnn,glognn} reveal that utilizing positive weights to aggregate neighbor node information mainly preserves the low-frequency information of nodes, which is inefficient in capturing the complex relations of node pairs on different types of graphs, such as homophily graphs and heterophily graphs. 
Hence, these methods develop signed weight-based strategies to aggregate different frequency information from nodes~\cite{ggcn, fagcn, glognn} or multi-hop neighborhoods~\cite{gprgnn}. 
Note that we focus on attributed graphs with no signed information in this paper.
We skip discussing recent works~\cite{sigraph1,sigraph2,sigraph3} that develop signed attention-based GNNs for signed graphs with positive and negative links since these methods require the signed information of edges as the model input.

Despite effectiveness, GNN-based methods only consider information of limited neighborhoods, which ignores information of nodes beyond local neighbors, leading to learning suboptimal node representations. 
Though GloGNN~\cite{glognn} claims to capture the information of global nodes, its aggregation coefficient matrix is forced to approximate the adjacency matrix with fixed propagation steps according to the objective function of GloGNN. 
It indicates that GloGNN mainly focuses on limited neighborhoods, inevitably overlooking information on long-range dependencies.
By comparison, our proposed \name leverages the signed self-attention mechanism that can carefully capture various frequency information of all nodes on the graph.

\subsection{Graph Transformer}
Recently, several efforts~\cite{graphormer,sat,nodeformer,nagphormer} have been made to generalize the Transformer architecture on graphs, called graph Transformers.
One line of the existing graph Transformers is to design various strategies to enable Transformer to capture the structural information of graphs.
Ying~\etal~\cite{graphormer} leverage the shortest path of node pairs as the structural information bias to strengthen the original attention matrix.
Dwivedi~\etal~\cite{gt} and Kreuzer~\etal~\cite{san} exploit the eigenvectors of graph Laplacian matrix as part of the input features to preserve the local topology information of nodes.
Bo~\etal~\cite{specformer} utilize the Transformer architecture to preserve both magnitudes and relative differences of all eigenvalues of the graph Laplacian.
Jain~\etal~\cite{graphtrans} and Chen~\etal~\cite{sat} develop hybrid neural network frameworks that first utilize GNN layers to extract the structural information of graphs and then leverage Transformer layers to learn node representations.

Another line of studies aims to develop efficient graph Transformers to reduce the training cost when handling large graphs.
A typical strategy is to introduce the linear Transformer module~\cite{gps,nodeformer}.
For instance, NodeFormer~\cite{nodeformer} replaces the original calculation of the attention matrix with a positive-definite kernel~\cite{performer} to achieve linear computational complexity.
Besides introducing various linear Transformer models, 
there are several studies~\cite{gophormer,ansgt,nagphormer} that generalize graph Transformers on large graphs through sampling node sequences from graphs.
Gophormer~\cite{gophormer} samples local neighbor nodes to construct the node sequences, while ANS-GT~\cite{ansgt} conducts a reinforcement learning-based strategy to adaptively sample nodes from different sampling strategies. 
Different from the above methods of sampling nodes as tokens to construct node sequences,
Chen~\etal~\cite{nagphormer} treat the neighborhoods of nodes as tokens and develop a novel module called Hop2Token that transforms multi-hop neighborhoods of nodes into tokens to construct the node sequences.

Existing graph Transformers utilize the attention matrix with all positive values to learn node representations, which has been proven as a smoothing operation on all nodes, preserving the low-frequency information of graphs~\cite{feta}.
Such a paradigm is inefficient in capturing complex node relations on different graphs.
Although Bastos~\etal~\cite{feta} develop a FeTA framework that enables the Transformer to capture different-frequency information of graphs, it is a hybrid model that mainly depends on spectral GNNs to enhance the modeling ability for graphs.
The main limitation of the Transformer architecture still remains: the self-attention module only captures low-frequency information of graphs. 

In this work, we propose a signed self-attention mechanism that generates signed attention values based on the semantic relevance of node pairs, which enables the Transformer to capture 
various frequency information of the node pairs adaptively.

\section{Preliminaries}

In this section, we first briefly introduce notations of attributed graphs, then introduce the Transformer architecture. 
In the end, we review the graph attention network and its variants expanded by signed attention.

\subsection{Notations}
Given an attributed graph $\mathcal{G}=(V,E)$ where $V$ is the node set and $E$ is the edge set, 
we have the corresponding adjacency matrix $\mathbf{A}\in \mathbb{R}^{n\times n}$ ($n$ is the number of nodes).
The normalized adjacency matrix is denoted as  $\hat{\mathbf{A}}= \tilde{\mathbf{D}}^{-1/2}\tilde{\mathbf{A}}\tilde{\mathbf{D}}^{-1/2}$,
where $\tilde{\mathbf{A}}=\mathbf{A} + \mathbf{I}$, $\mathbf{I}\in \mathbb{R}^{n\times n}$ is the identity matrix, and $\tilde{\mathbf{D}}$ is the corresponding diagonal degree matrix of $\tilde{\mathbf{A}}$.
$\mathbf{X}\in \mathbb{R}^{n\times d}$ represents the attribute feature matrix of nodes, where $d$ is the dimension of the feature vector.

\subsection{Transformer}
Transformer is an advanced deep learning architecture. 
Each Transformer layer contains two core modules, \ie multi-head self-attention (MHA) and feed-forward networks (FFN).

Given the input feature matrix $\mathbf{H} \in \mathbb{R}^{n\times d}$, the output through single head self-attention is calculated as:
\begin{equation}
    \mathrm{Attention}(\mathbf{Q},\mathbf{K},\mathbf{V}) = \mathit{softmax}(\frac{\mathbf{Q}\mathbf{K}^{\mathrm{T}}}{\sqrt{d_k}})\mathbf{V},
    \label{sgatt}
\end{equation}
where $\mathbf{Q}=\mathbf{H}\mathbf{W}^{Q}$, 
$\mathbf{K}=\mathbf{H}\mathbf{W}^{K}$,
$\mathbf{V}=\mathbf{H}\mathbf{W}^{V}$.
$\mathbf{W}^{Q}\in \mathbb{R}^{d\times d_k}$, 
$\mathbf{W}^{K}\in \mathbb{R}^{d\times d_k}$,
$\mathbf{W}^{V}\in \mathbb{R}^{d\times d_v}$ are projection matrices.
And the output of MHA is calculated as:
\begin{equation}
    \mathrm{MHA}(\mathbf{Q},\mathbf{K},\mathbf{V}) = (head_1||head_2||\cdots||head_h)\mathbf{W}^{O},
    \label{mha}
\end{equation}
where $head_i = \mathrm{Attention}(\mathbf{H}\mathbf{W}^{Q}_{i},\mathbf{H}\mathbf{W}^{K}_{i},\mathbf{H}\mathbf{W}^{V}_{i}), i\in \{1,\cdots,h \}$.
$\mathbf{W}^{O}\in \mathbb{R}^{d_v \times d}$ represents the learnable parameter matrix.

The FFN contains two linear layers and a nonlinear activation function, which is described as follows:
\begin{equation}
    \mathrm{FFN}(\mathbf{H}) = \mathrm{Linear}(\sigma(\mathrm{Linear(\mathbf{H})})),
    \label{ffn}
\end{equation}
where $\mathrm{Linear}(\cdot)$ denotes the linear layer and $\sigma(\cdot)$ denotes the activation function.

\subsection{Graph Attention Network}\label{sgat}
Graph attention network (GAT)~\cite{gat} introduces the attention mechanism into GNN to aggregate the information of immediate neighbors adaptively.
For a node $v_i$, the attention weight of the neighbor node $v_j$ in GAT is calculated as follows:
\begin{equation}
\alpha_{v_i v_j}=\frac{\exp (\sigma(\mathbf{a}^{\mathrm{T}}[\mathbf{W} \mathbf{H}_{v_i} \| \mathbf{W} \mathbf{H}_{v_j}]))}{\sum_{v_k \in \mathcal{N}_{v_i}} \exp \left(\sigma\left(\mathbf{a}^{\mathrm{T}}\left[\mathbf{W} \mathbf{H}_{v_i} \| \mathbf{W} \mathbf{H}_{v_k}\right]\right)\right)},
\label{orgat}
\end{equation}
where $\mathbf{H}_{v_i}$ and $\mathbf{H}_{v_j}$ are the representations of $v_i$ and $v_j$.
$\mathbf{W}$ and $\mathbf{a}$ are the learnable projection matrix and projection vector, respectively. $\sigma(\cdot)$ denotes the nonlinearity activation function.
$\mathcal{N}_{v_i}$ represents the immediate neighbor set of $v_i$.

According to \autoref{orgat}, the attention weight $\alpha_{v_i v_j}\in (0,1]$ generated by GAT are always positive, which is inefficient in capturing various frequency information of connected nodes~\cite{fagcn}.
This situation stimulates the signed attention-based GNNs~\cite{fagcn,sign1,sign2}.
These methods share a similar idea, utilizing the $\mathit{tanh}(\cdot)$ to produce the attention weights.

We take FAGCN~\cite{fagcn} as the example. The attention weight in FAGCN is calculated as follows:
\begin{equation}
\alpha_{v_i v_j}^{\prime}=\mathit{tanh}(\mathbf{a}^{\mathrm{T}}[\mathbf{H}_{v_i} \| \mathbf{H}_{v_j}]).
\label{faatt}
\end{equation}
Compared to \autoref{orgat}, $\mathit{tanh}(\cdot)$ forces the attention weight $\alpha_{v_i v_j}^{\prime}$ into $(-1,1)$.
Bo \etal~\cite{fagcn} have proved that FAGCN captures the low-frequency information of the node pair $(v_i,v_j)$, when $\alpha_{v_i v_j}^{\prime} > 0$. And $\alpha_{v_i v_j}^{\prime} < 0$ indicates that FAGCN preserves the high-frequency information of $(v_i,v_j)$.
\section{Methodology}

In this section, we first discuss the motivation of our method.
Then, we introduce our signed self-attention mechanism, \saname, which is tailored for Transformer.
Finally, we detail the architecture of \name that consists of a new graph Transformer layer built on \saname and \sffn, an extension of FFN 
with structural bias.

\subsection{Motivation}
There are mainly two types of frequency information on graphs: low-frequency information and high-frequency information \cite{fagcn}.

Specifically, the low-frequency information for each connected node pair mainly preserves the commonality in node representations. 
In contrast, the high-frequency information mainly captures the difference in node representations.
When conducting Transformer on graphs, we can rewrite \autoref{sgatt} from the perspective of aggregation operation:
\begin{equation}
    \mathbf{H}^{\prime}_{v_i} = \sum_{j=1}^{n} \mathbf{M}_{v_i, v_j} \cdot \mathbf{V}_{v_j}, ~~
    \mathbf{M}_{v_i, v_j} = 
    \frac{\mathrm{exp}(\mathbf{Q}_{v_i}
    \mathbf{K}_{v_j}^{\mathrm{T}})}
    {\sum_{k=1}^{n}\mathrm{exp}(\mathbf{Q}_{v_i}
    \mathbf{K}_{v_k}^{\mathrm{T}})},
    \label{agg_att}
\end{equation}
where $\mathbf{M}_{v_i, v_j}$ denotes the attention weight from $v_j$ to $v_i$.
We omit the scaling operation and the multi-head strategy for simplified expression.

Based on \autoref{agg_att}, we can observe that the self-attention mechanism 
is equivalent to conduct a single 
aggregation operation across all the nodes.
Due to the characteristic of the exponential function, all attention weights of the attention matrix $\mathbf{M}$ are positive, which mainly preserves low-frequency information of the node pairs~\cite{fagcn}.
Moreover, a recent study~\cite{feta} also reveals that the self-attention mechanism of Transformer is only effective in learning low-frequency information from the perspective of spectral space.

Since the self-attention mechanism regards the input graph as fully connected, each node must aggregate information from all nodes on the graph.
On the one hand, such characteristic enables the model to easily capture long-range dependencies on graphs~\cite{graphtrans}, which is the biggest advantage compared to GNNs.
On the other hand, such aggregation operation also involves information from many irrelevant nodes, \eg two nodes belonging to different labels.
In this situation, the high-frequency information is crucial for learning informative node representations, \ie preserving the difference in node representations.
Unfortunately, as mentioned before, the self-attention mechanism fails to capture high-frequency information of the node pairs effectively.
This phenomenon motivates us to design a new self-attention mechanism that enables Transformer to capture various frequency information on graphs.
  
\subsection{Signed Self-Attention Mechanism}
To imbue the self-attention mechanism with the ability to capture various frequency information, a feasible solution is to generate signed attention values according to the input node pairs, where positive values preserve low-frequency information and negative values capture high-frequency information~\cite{fagcn}.
As discussed before, existing signed attention mechanisms built on the GAT backbone leverage $\mathit{tanh}(\cdot)$ to generate the signed attention values.
This strategy is not suitable for Transformer-based methods due to the lack of normalization operation.
This situation implies that a new signed attention mechanism tailored for Transformer is necessary and crucial.

In this paper, we design a new signed self-attention mechanism (\saname) based on the original self-attention calculation.
Specifically, for each node pair $(v_i, v_j)$, the attention weight of \saname is calculated as:
\begin{equation}
    \mathbf{M}^{S}_{v_i, v_j} = \mathrm{sgn}(\mathbf{Q}_{v_i}\mathbf{K}_{v_j}^{\mathrm{T}}) \cdot
    \frac{\mathrm{exp}(|\mathbf{Q}_{v_i}
    \mathbf{K}_{v_j}^{\mathrm{T}}|)}
    {\sum_{k=1}^{n}\mathrm{exp}(|\mathbf{Q}_{v_i}
    \mathbf{K}_{v_k}^{\mathrm{T}}|)}.    
    \label{sa_att}
\end{equation}
Here $\mathrm{sgn}(\cdot)$ denotes the sign function, and 
$|\cdot|$ denotes the absolute value.
The rationale of \autoref{sa_att} is to separate the sign information of attention values before normalization.
The separated sign information could be regarded as the semantic relevance of each node pair.
Ideally, when $v_i$ and $v_j$ exhibit dissimilar representation vectors, $\mathrm{sgn}(\cdot)$ will produce a negative signal, which forces the node representations to become more dissimilar, capturing the high-frequency information between the input nodes.
Otherwise, $\mathrm{sgn}(\cdot)$ will produce a positive signal, which forces the node representations to be similar, preserving the low-frequency information.

The advantages of \saname compared to the original attention mechanism are two-fold:
(1) By separating the sign information of attention weights, \saname can adaptively generate the signed attention weights based on the semantic relevance of nodes, enabling the attention matrix to preserve different frequency information of graphs. 
(2) \saname completely preserves the influences of both positive and negative signals on the target node.
Since \saname leverages the absolute value of the dot product as the input of the exponential function, the influence of negative signals can be well preserved.

\subsection{Architecture of \name}
Here, we detail the architecture of the proposed \name, which consists of a new graph Transformer layer, called \name layer.
Each \name layer contains two core components, \saname module and \sffn module.
The former leverages the proposed \saname to capture various frequency information from the whole input graph.
While the latter extends the original FFN module in Transformer by introducing neighborhood structural bias to preserve the local topology information.

\textit{\textbf{\saname module.}}
We replace the original self-attention mechanism in Transformer with \saname.
\autoref{sa_att} depicts the calculation of the single-head attention for \saname.
After obtaining the attention values, for a given node $v_i$, we update its representation $\mathbf{H}_{v_i}^{\prime}$ as follows:
\begin{equation}
    \mathbf{H}_{v_i}^{\prime} = \sum_{j=1}^{n} \mathbf{M}_{v_i, v_j}^{S} \cdot \mathbf{V}_{v_j}.
    \label{agg_sum}
\end{equation}
The above calculation is easily extended to the multi-head attention mode via \autoref{mha}, we omit this process for simplified description.

\textit{\textbf{\sffn module.}}
Although \saname can capture various frequency information from the whole graph, it naturally overlooks graph topology features since it regards the input graph as fully connected.
Previous studies of GNNs~\cite{fagcn,gprgnn,gcn,gdc} have  demonstrated that local topology features of nodes are also essential for learning informative node representations. 
Existing graph Transformers~\cite{nodeformer,gps,sat} utilize the learnable bias or the standard GNN layer to aggregate information of immediate neighbor nodes for enabling the Transformer to preserve local topology information.
Nevertheless, these strategies could make the model hard to effectively capture local topology information beyond immediate neighbors.

Thereby, we propose \sffn that integrates the neighborhood structural bias into the original FFN module to preserve the local topology information. 
The representation of $v_i$ learned by \sffn is described as follows:
\begin{equation}
    \mathbf{H}_{v_i}^{\prime\prime} = \mathrm{Linear}\left(\sigma
    \left(\sum_{v_j \in \mathcal{N}_{v_i}^{k}} \mathbf{W}^{S}_{v_i,v_j} \cdot \mathrm{Linear}(\mathbf{H}_{v_j}^{\prime})\right)\right),
    \label{sffn}
\end{equation}
where $\mathbf{W}^{S}_{v_i,v_j}$ denotes the structural bias of the node pair $(v_i, v_j)$, preserving the influence of $v_j$ on $v_i$ from the perspective of the graph topology structure.
$\mathcal{N}_{v_i}^{k}$ denotes the $k$-hop neighborhood node set of $v_i$. 
Inspired by recent advanced GNNs~\cite{sgc,rlp,gdc}, we utilize the propagation method to calculate the structural bias. 
Specifically, for each $k$-hop neighborhood node $v_j$ of $v_i$, $\mathbf{W}^{S}_{v_i,v_j}$ is calculated by: 
\begin{equation}
    \mathbf{W}^{S}_{v_i,v_j}=\bar{\mathbf{A}}_{v_i,v_j}, ~ \bar{\mathbf{A}} =\hat{\mathbf{A}}^{k}.
    \label{graphbias}
\end{equation}

Compared to existing graph Transformers~\cite{nodeformer,gps,transformer}, 
\sffn flexibly captures the local topology information of $k$-hop neighborhood without introducing additional model parameters or calculation operation during the model training since the structural bias $\mathbf{W}^{S}$ could be pre-computed with the non-parametric propagation method before the training stage.

\begin{figure}[t]
\centering
\includegraphics[width=8.5cm]{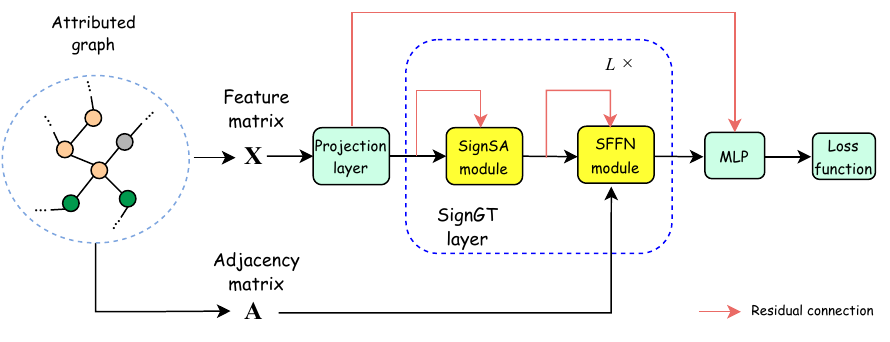}
\caption{
The overall framework of \name.
}
\label{fig:fw}
\end{figure}

\textit{\textbf{Overall framework.}}
\autoref{fig:fw} illustrates the overall framework of \name.
Specifically, we first utilize the projection layer to obtain the hidden representations of nodes for dimension reduction since the dimension of raw attribute features on some graphs could be very large.
Then, we leverage the neural network backbone consisting of the designed layers to generate the final representation of nodes.
The residual connection technique is adopted in practice.

\subsection{\name for Graph Mining Tasks}
Through several \name layers, we could obtain the final node representations $\mathbf{H}^{out}\in \mathbb{R}^{n \times d_{out}}$. 
The learned node representations are used for various downstream graph mining tasks via different loss functions.
In this way, \name is capable of dealing with various graph mining tasks, including node-level and graph-level tasks. 
For node-level tasks, \eg, node classification, we can directly feed the learned node representations to the MLP.
For graph-level tasks, \eg, graph classification, we first leverage a pooling function to obtain the representation of the entire graph and then use an MLP to predict graph labels.

\section{Experiments}

In this section, we evaluate the performance of \name on node-level and graph-level classification tasks, namely node classification and graph classification. 
We first briefly introduce the experimental setup for each task, including datasets and baselines.
The implementation details, such as statistics of datasets, software environment, and hyper-parameter settings, are reported in the Appendix.
Then, we report and analyze the performances of all models on these tasks. 
Finally, we conduct a series of analytical experiments, including ablation study, parameter sensitivity analysis, and visualization, to understand \name more deeply.

\begin{table*}[t]
\caption{Comparison of all models in terms of mean accuracy $\pm$ stdev (\%). 
	The best results appear in \textbf{bold}. The second results appear in \underline{underline}.}
\label{tab:dere}
\centering
\renewcommand\arraystretch{1.0}
\scalebox{0.84}{
\begin{tabular}{lcccccccccccc}
\toprule
Dataset& Photo & ACM & Cora & Pubmed & Computer & CoraFull & BlogCatalog & UAI2010 & Flickr &Chameleon &Actor &Squirrel \\
$\mathcal{H}$& 0.83 & 0.82 & 0.81 & 0.80 & 0.78 & 0.57& 0.40 & 0.36 & 0.24  &0.24& 0.22 & 0.21 \\ \hline
GCN& 94.76\tiny{$\pm$0.19} & 92.33\tiny{$\pm$0.38} & 87.29\tiny{$\pm$0.37} & 86.66\tiny{$\pm$0.14} & 89.53\tiny{$\pm$0.41} & 61.88\tiny{$\pm$0.19}        & 93.82\tiny{$\pm$0.36} & 74.48\tiny{$\pm$0.86} & 86.37\tiny{$\pm$0.40} & 59.62\tiny{$\pm$2.01}      & 35.45\tiny{$\pm$0.42} & 45.63\tiny{$\pm$1.78}  \\
GAT& 94.98\tiny{$\pm$0.24} & 94.24\tiny{$\pm$0.43} & 88.96\tiny{$\pm$0.29} & 86.45\tiny{$\pm$0.13} & 90.78\tiny{$\pm$0.13} & 64.62\tiny{$\pm$0.23}        & 92.62\tiny{$\pm$0.31} & 75.22\tiny{$\pm$0.52} & 85.73\tiny{$\pm$0.62}         & 49.98\tiny{$\pm$1.92} & 30.18\tiny{$\pm$0.21} & 36.58\tiny{$\pm$2.11}    \\
SGC& 95.11\tiny{$\pm$0.09} & 93.15\tiny{$\pm$0.22} & 87.66\tiny{$\pm$0.28} & 86.42\tiny{$\pm$0.16} &  90.22\tiny{$\pm$0.21} & 63.16\tiny{$\pm$0.28}        & 93.12\tiny{$\pm$0.39} & 72.16\tiny{$\pm$0.63}  &  86.79\tiny{$\pm$0.46}         & 51.49\tiny{$\pm$1.83} & 29.97\tiny{$\pm$0.26}  & 36.74\tiny{$\pm$2.56}  \\ \hline
GPRGNN& \underline{95.49\tiny{$\pm$0.14}} & 94.42\tiny{$\pm$0.49} & \underline{88.97\tiny{$\pm$0.33}} & \underline{89.38\tiny{$\pm$0.32}} & 89.32\tiny{$\pm$0.29}   & 67.56\tiny{$\pm$0.35} & 95.42\tiny{$\pm$0.26} & 76.19\tiny{$\pm$0.82} & 90.66\tiny{$\pm$0.68} & 69.85\tiny{$\pm$2.12} & 37.86\tiny{$\pm$0.19}     & 54.37\tiny{$\pm$2.18}   \\
FAGCN& 94.56\tiny{$\pm$0.29} & 94.87\tiny{$\pm$0.59} & 88.77\tiny{$\pm$0.91} & 89.14\tiny{$\pm$0.26} & 87.11\tiny{$\pm$0.61} & \underline{70.26\tiny{$\pm$0.67}}        & 96.05\tiny{$\pm$0.43} & 74.06\tiny{$\pm$0.65} & \underline{92.19\tiny{$\pm$0.32}}         & 69.67\tiny{$\pm$1.75} & 37.63\tiny{$\pm$0.61} & \underline{63.21\tiny{$\pm$2.97}}  \\
GloGNN& 95.17\tiny{$\pm$0.29} & 94.07\tiny{$\pm$0.46} & 88.57\tiny{$\pm$0.42} & 88.58\tiny{$\pm$0.41}  & 89.12\tiny{$\pm$0.33}        & 67.35\tiny{$\pm$0.25}        & 93.94\tiny{$\pm$0.38}  & \underline{76.52\tiny{$\pm$0.96}}  & 91.14\tiny{$\pm$0.42}         & \underline{74.16\tiny{$\pm$1.42}} & \textbf{39.59\tiny{$\pm$0.29}}      & 59.75\tiny{$\pm$1.81}  \\ \hline
NodeFormer& 95.27\tiny{$\pm$0.22} & \underline{94.93\tiny{$\pm$0.35}} & 87.32\tiny{$\pm$0.92}    & 89.24\tiny{$\pm$0.23}  & \underline{91.12\tiny{$\pm$0.43}}        & 61.82\tiny{$\pm$0.81}        & 94.33\tiny{$\pm$0.23} & 75.84\tiny{$\pm$0.86} & 90.97\tiny{$\pm$0.44}         & 56.34\tiny{$\pm$1.11}  & 34.62\tiny{$\pm$0.82}      & 43.42\tiny{$\pm$0.62}  \\
Specformer& 94.51\tiny{$\pm$0.27} & 94.54\tiny{$\pm$0.43} & 88.41\tiny{$\pm$0.43}& 89.19\tiny{$\pm$0.28}      & 91.02\tiny{$\pm$0.26}        & 66.58\tiny{$\pm$0.95}        & \underline{96.12\tiny{$\pm$0.22}} & 75.61\tiny{$\pm$0.62}       & 92.05\tiny{$\pm$0.38}         & 73.31\tiny{$\pm$2.01} & 38.26\tiny{$\pm$0.52}     & 62.15\tiny{$\pm$1.51}   \\ \hline
\name & \textbf{95.68\tiny{$\pm$0.36}} & \textbf{95.45\tiny{$\pm$0.49}} & \textbf{89.34\tiny{$\pm$0.35}} & \textbf{89.64\tiny{$\pm$0.45}} & \textbf{91.71\tiny{$\pm$0.44}} & \textbf{70.73\tiny{$\pm$0.46}} & \textbf{96.98\tiny{$\pm$0.35}} & \textbf{78.72\tiny{$\pm$0.69}} & \textbf{92.91\tiny{$\pm$0.36}} & \textbf{74.31\tiny{$\pm$1.24}} & \underline{38.65\tiny{$\pm$0.32}} & \textbf{64.25\tiny{$\pm$1.48}}     \\     
 \toprule
\end{tabular}
}
\end{table*}

\begin{table}[t]
\caption{Comparison of all models in terms of mean accuracy $\pm$ stdev (\%). 
	The best results appear in \textbf{bold}. The second results appear in \underline{underline}.}
\label{tab:gc_res}
\centering
\renewcommand\arraystretch{1.0}
\scalebox{0.84}{
\begin{tabular}{lccccc}
\toprule
Dataset& NCI1 & NCI109 & Mutag. & FRANK. & COLLAB  \\
\hline
GCN & 79.68\tiny{$\pm$2.05} & 78.05\tiny{$\pm$1.59} & 79.81\tiny{$\pm$1.58} & 62.71\tiny{$\pm$0.79} & 71.92\tiny{$\pm$3.24}  \\

GAT & 79.88\tiny{$\pm$0.88} & 79.93\tiny{$\pm$1.52} & 78.89\tiny{$\pm$2.05} & 63.11\tiny{$\pm$0.61} & 75.80\tiny{$\pm$1.60} \\

GraphSAGE & 
78.98\tiny{$\pm$1.84} & 
77.27\tiny{$\pm$1.66} & 
78.75\tiny{$\pm$1.18} & 
62.88\tiny{$\pm$0.55} & 
79.70\tiny{$\pm$1.70}\\ \hline

Set2Set & 
68.62\tiny{$\pm$1.90} & 
69.88\tiny{$\pm$1.20} & 
80.84\tiny{$\pm$0.67} & 
61.46\tiny{$\pm$0.47} & 
65.34\tiny{$\pm$0.44} \\
SAGPool$_h$ & 
67.55\tiny{$\pm$1.03} & 
67.91\tiny{$\pm$0.85} & 
79.72\tiny{$\pm$0.79} & 
61.73\tiny{$\pm$0.76} & 
73.08\tiny{$\pm$1.31} \\
SortPool & 
73.42\tiny{$\pm$1.12} & 
73.53\tiny{$\pm$0.91} & 
80.43\tiny{$\pm$1.12} & 
63.44\tiny{$\pm$0.65} & 
71.18\tiny{$\pm$2.12} \\ \hline
Transformer & 
68.47\tiny{$\pm$1.78} & 
70.24\tiny{$\pm$1.29} & 
74.25\tiny{$\pm$0.69} & 
61.04\tiny{$\pm$0.93} & 
69.57\tiny{$\pm$3.22} \\
GT & 
80.15\tiny{$\pm$2.04} & 
78.94\tiny{$\pm$1.15} & 
80.79\tiny{$\pm$0.98} & 
68.54\tiny{$\pm$0.61} & 
79.63\tiny{$\pm$1.02} \\
GraphTrans & 
81.27\tiny{$\pm$1.90} & 
79.20\tiny{$\pm$2.20} & 
81.54\tiny{$\pm$0.91} & 
\underline{70.22\tiny{$\pm$0.77}} & 
79.81\tiny{$\pm$0.84} \\
SAT & 
80.69\tiny{$\pm$1.55} & 
79.06\tiny{$\pm$0.89} & 
\underline{81.62\tiny{$\pm$1.03}} & 
70.18\tiny{$\pm$0.57} & 
80.05\tiny{$\pm$0.55}\\
Graphormer & 
\underline{81.44\tiny{$\pm$0.57}} & 
\underline{79.91\tiny{$\pm$1.14}} & 
81.15\tiny{$\pm$1.25} & 
69.72\tiny{$\pm$1.19} & 
\underline{81.80\tiny{$\pm$2.24}} \\ \hline

\name & \textbf{83.42\tiny{$\pm$0.72}} & \textbf{80.48\tiny{$\pm$1.25}} & \textbf{83.14\tiny{$\pm$0.56}} & \textbf{70.93\tiny{$\pm$0.39}} & \textbf{82.40\tiny{$\pm$0.14}} \\     
 \toprule
\end{tabular}
}
\end{table}


\subsection{Node Classification}
The goal of node classification is to predict the labels of nodes.
We first introduce the datasets and baselines for the node classification task.
Then we report the model performance on this task.

\textbf{Datasets.} 
We adopt 12 benchmark datasets containing homophily and heterophily graphs derived from diverse domains. 
\texttt{Cora}, \texttt{Pubmed}, \texttt{ACM}, \texttt{UAI2010}, \texttt{CoraFull} are citation networks.
\texttt{BlogCatalog} and \texttt{Flick} are social networks.
\texttt{Computer} and \texttt{Photo} are co-purchase networks.
\texttt{Actor} is the actor co-occurrence network.
\texttt{Squirrel} and \texttt{Chameleon} are Wikipedia networks. 
\texttt{ACM}, \texttt{UAI2010}, \texttt{CoraFull}, \texttt{BlogCatalog} and \texttt{Flick} are from~\cite{amgcn}, and others are from~\cite{gprgnn}.

For each dataset, We follow the 
metric in~\cite{glognn} to measure the level of homophily:
\begin{equation}
    \mathcal{H}(\mathcal{G}) =  \frac{|\{e_{i,j} |(v_i,v_j)\in E, \mathbf{Y}_{v_i}=\mathbf{Y}_{v_j}\}|}{|E|},
\end{equation}
where $\mathbf{Y}_{v_i}$ denotes the label of node ${v_i}$.
Low $\mathcal{H}(\mathcal{G})$ means the graph exhibits strong heterophily.
The detailed information of datasets is reported in Appendix~\ref{app_dataset}.
Following the settings of previous studies~\cite{fagcn,gprgnn,specformer}, we randomly choose 60\% of each label as the training set, 20\% as the validation set, and the rest as the test set.

\textbf{Baselines.}
We select eight representative baselines from two categories: 
GNN-based methods and graph Transformer-based methods.
For GNN-based methods, we choose GCN~\cite{gcn}, GAT~\cite{gat}, SGC \cite{sgc}, GPRGNN~\cite{gprgnn}, FAGCN~\cite{fagcn} and GloGNN~\cite{glognn}.
The last three methods develop various signed weight-based aggregation strategies to learn node representations for both homophily and heterophily graphs.
For graph Transformer-based methods, we select NodeFormer~\cite{nodeformer} and Specformer~\cite{specformer}, which are recent graph Transformers designed for the node classification task.

\textbf{Performance.}
To evaluate the model performance on node classification,
we run each model ten times with different random seeds and then calculate the average accuracy values. 
The experimental results are reported in \autoref{tab:dere}.
We can observe that \name obtains the best performance on most benchmark datasets, except on \texttt{Actor} in which it gains the second best.
The result demonstrates the effectiveness of \name for node classification on diverse graphs. 
Specifically, \name beats NodeFormer and Specformer, indicating that the designs of \name are more suitable for learning informative node representations from various graphs than recent advanced graph Transformers. 

In addition, we can observe that methods allowing signed weights to aggregate information of neighbors can achieve better performance than others whose aggregation weights are always positive on most graphs, especially on heterophily graphs since high-frequency information is essential in these graphs.
For instance, FAGCN obtains strong performance on \texttt{Corafull}, \texttt{BlogCatalog}, \texttt{Flickr}, and \texttt{Squirrel}.
This phenomenon exhibits that preserving different frequency information is more beneficial to learning node representations.
However, restricted by the message-passing mechanism, GNNs only preserve different-frequency information in local neighborhoods.
In contrast, \name can capture various frequency information on the whole graph via the proposed \saname, and outperforms these advanced GNNs on almost all datasets, indicating that capturing global information is beneficial to enhancing the modeling capacity for graph representation learning.

\begin{table*}[t]
\caption{Performances of \name and its variants on the node classification task.}
\label{tab:attre}
\centering
\renewcommand\arraystretch{1.0}
\scalebox{0.84}{
\begin{tabular}{lcccccccccccc}
\toprule
Dataset& Photo & ACM & Cora & Pubmed & Computer & CoraFull & BlogCatalog & UAI2010 & Flickr &Chameleon &Actor &Squirrel \\
 \hline
\name-O & 94.44& 94.01& 77.96& 88.76& 90.15& 69.37& 96.22& 77.89& 92.11& 43.22& 34.67& 36.26\\
\name-T & 40.94& 81.32& 37.22& 88.59& 37.42& 48.07& 41.87& 59.84& 46.01& 41.92& 35.86& 31.41\\
\name   & 95.68& 95.45& 89.34& 89.64& 91.71& 70.73& 96.98& 78.72& 92.91& 74.31& 38.65& 64.25\\  \hline
Gain    & +1.24& +1.44&+11.38& +0.88& +1.56& +1.36& +0.76& +0.83& +0.80&+31.09& +3.98&+27.99\\
 \toprule
\end{tabular}
}
\end{table*}
\begin{figure*}[t]
\centering
\includegraphics[width=14cm]{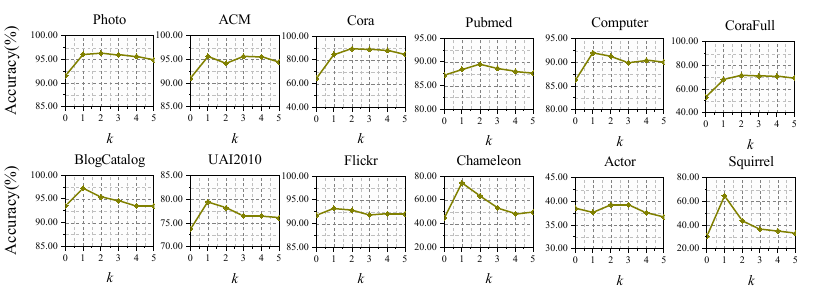}
\caption{
The influence of different neighborhood ranges on model performance.
}
\label{fig:k}
\end{figure*}

\subsection{Graph Classification}
Different from node classification, graph classification aims to predict the label of the whole given graph. 
Similar to node classification, we first introduce the adopted datasets and baselines, then report the model performance.

\textbf{Datasets.} 
We utilize five widely used benchmarks for graph classification.
Specifically, we select \texttt{NCI1}~\cite{nci}, \texttt{NCI109}~\cite{nci}, \texttt{COLLAB} \cite{collab}, \texttt{FRANKENSTEIN} \cite{frank} and \texttt{Mutagenicity}~\cite{mutag} from TUDatasets \cite{tudataset}. 
Among them, \texttt{NCI1}, \texttt{NCI109}, \texttt{Mutagenicity} and \texttt{FRANKENSTEIN} are molecule graphs, while \texttt{COLLAB} is comprised of various social networks.
The detailed information of datasets is reported in Appendix~\ref{app_dataset}.
Following the setup in \cite{nci, graphtrans}, we randomly split the dataset into training, validation and test sets by a ratio of $8:1:1$.

\textbf{Baselines.}
For the graph classification task, we take three kinds of popular representative methods as the baselines: (\uppercase\expandafter{\romannumeral1})
popular GNN-based methods, including GCN~\cite{gcn}, GAT~\cite{gat} and GraphSage~\cite{graphsage}. 
(\uppercase\expandafter{\romannumeral2}) graph pooling methods only for graph classification tasks, including Set2Set~\cite{set2set}, SAGpool~\cite{sagpool} and SortPool~\cite{sortpool}.
(\uppercase\expandafter{\romannumeral3}) graph Transformer models, including the original Transformer~\cite{transformer}, GT~\cite{gt}, GraphTrans~\cite{graphtrans},  SAT~\cite{sat} and Graphormer~\cite{graphormer}. 

\textbf{Performance.} 
We also run each model ten times on each dataset with different random seeds and report the average accuracy with standard deviation. 
The experimental results are summarized in Table \ref{tab:gc_res}. 
Generally speaking, \name consistently outperforms all the baselines, demonstrating the superiority of our model on the graph classification task.

Specifically, \name improves the performance by $1.98\%$, $0.57\%$, $1.52\%$, $0.71\%$ and $0.60\%$ over the best baselines on \texttt{NCI1}, \texttt{NCI109}, \texttt{Mutagenicity}, \texttt{FRANKENSTEIN} and \texttt{COLLAB}, respectively. 
In particular, we can observe that the performance of \name exceeds its graph Transformer counterparts, illustrating the efficacy of our new attention mechanism. 
Moreover, compared with GNN-based methods that only capture local information, our \name can efficiently encode long-range dependencies, thus obtaining better performance. 
These observations clearly illuminate the significance of capturing global information on different frequencies when learning the graph representations.

\begin{figure*} [ht]
	\centering
	\subfloat[GPRGNN]{
		\includegraphics[scale=0.15]{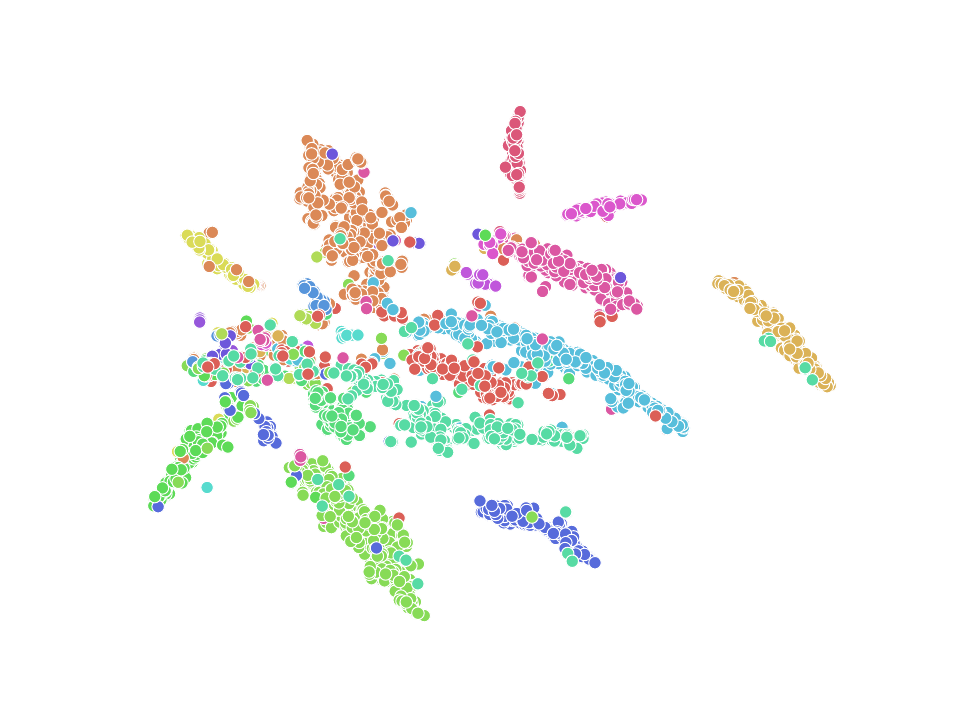}}
	\subfloat[FAGCN]{
		\includegraphics[scale=0.15]{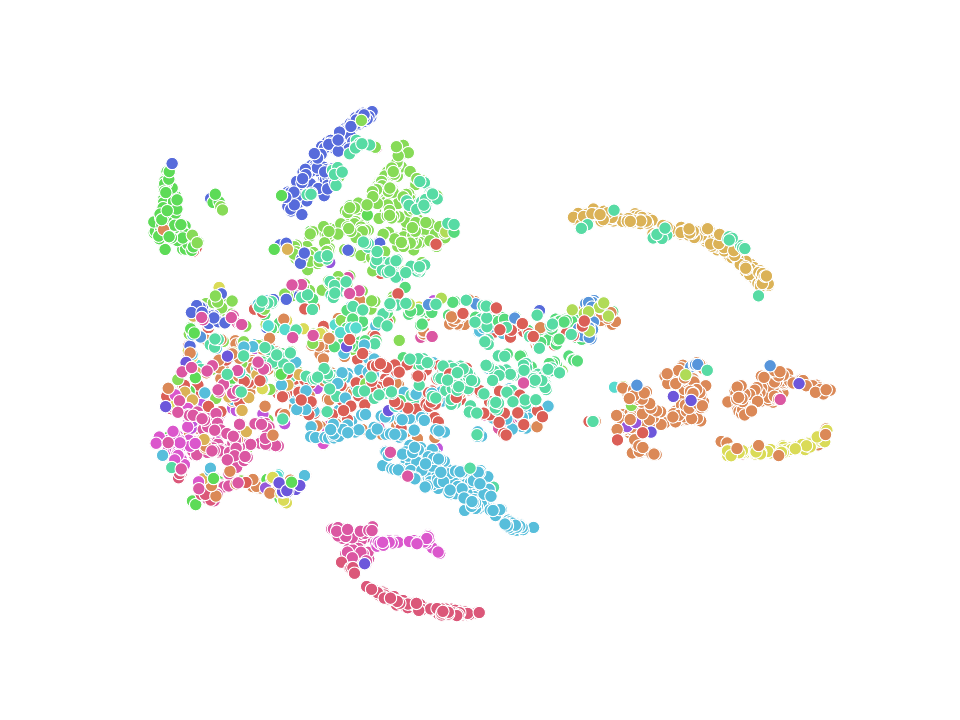}}
	\subfloat[GloGNN]{
		\includegraphics[scale=0.15]{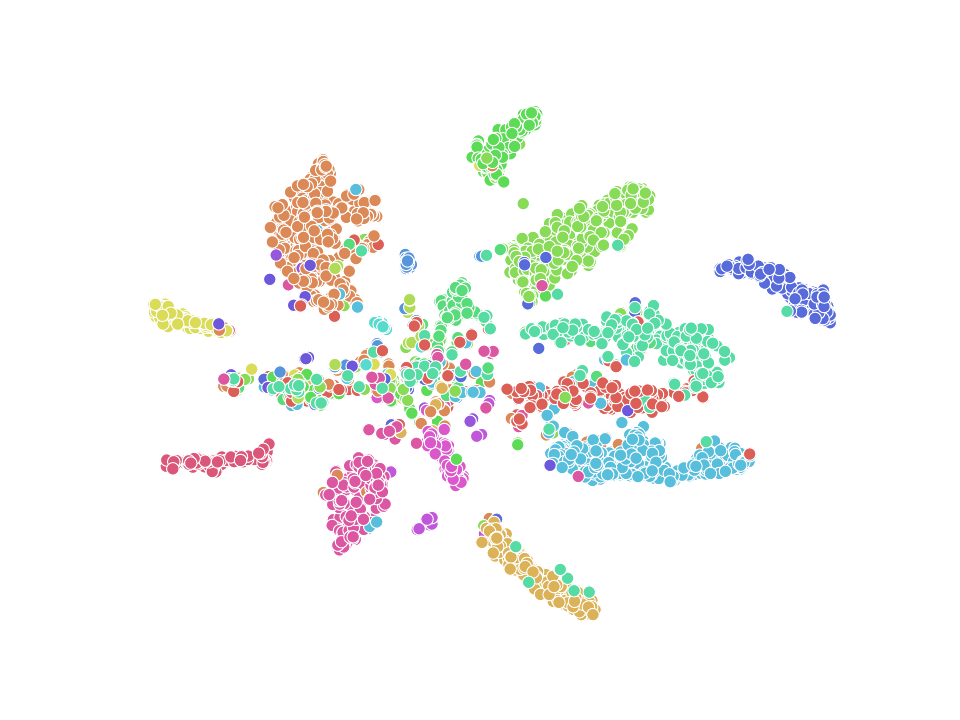}}
	\subfloat[NodeFormer]{
		\includegraphics[scale=0.15]{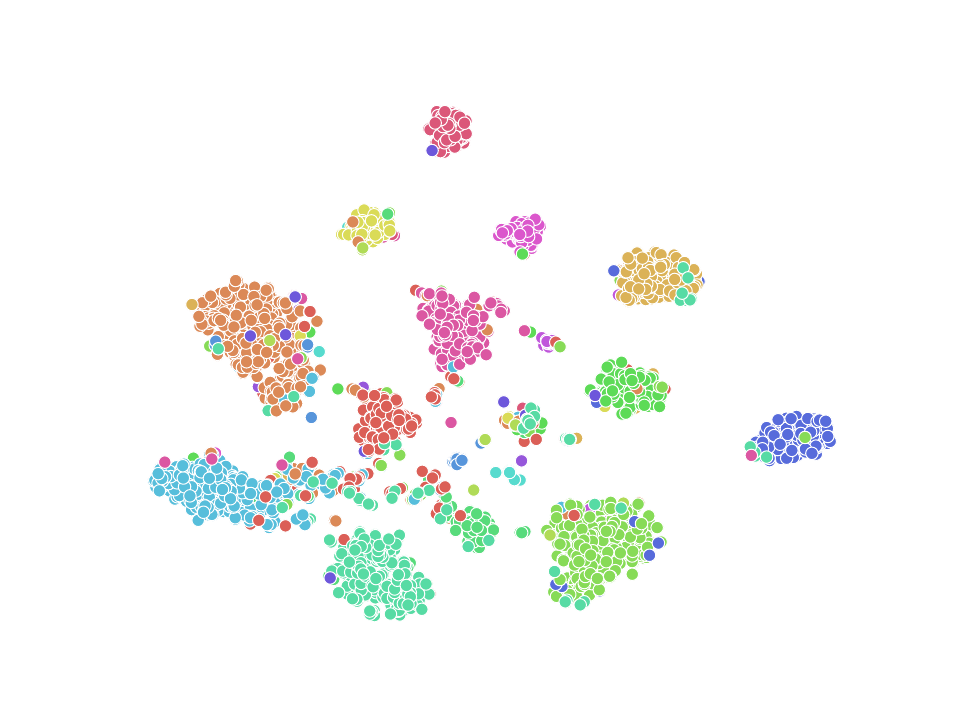} }
	\subfloat[Specformer]{
		\includegraphics[scale=0.15]{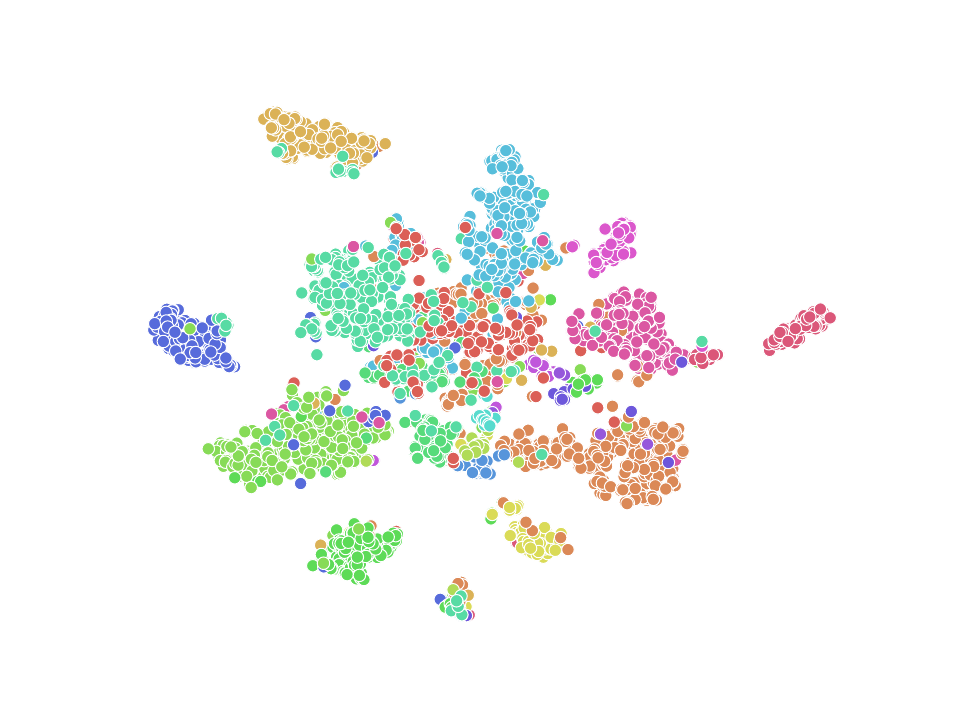}}
	\subfloat[\name]{
		\includegraphics[scale=0.15]{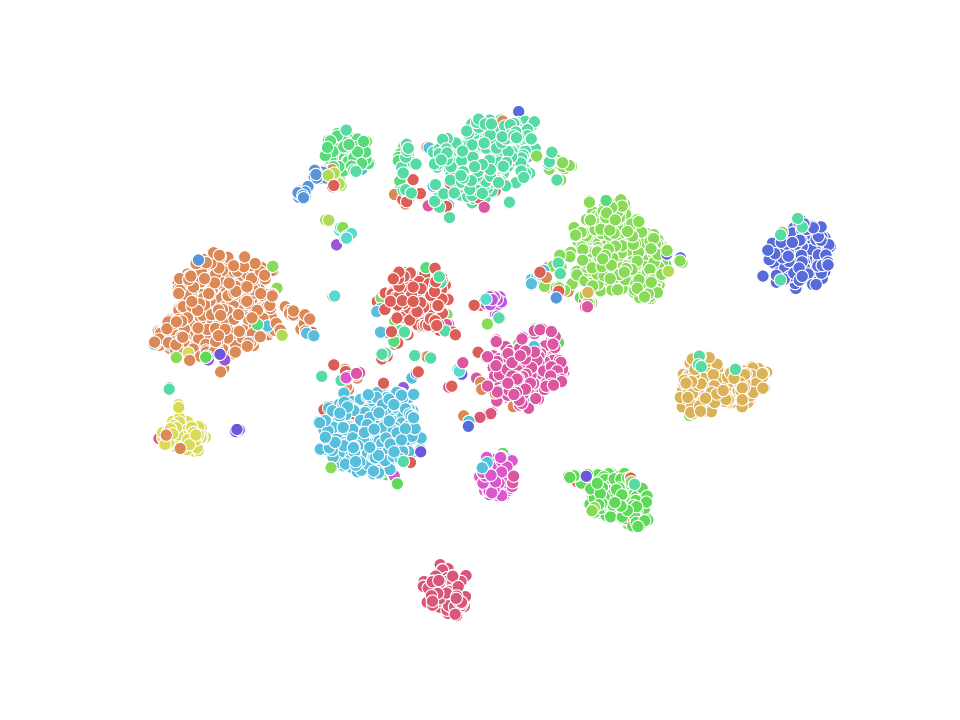}}
	\caption{Visualization of node representations learned by each model on \texttt{UAI2010}. Colors represent labels of nodes.}
	\label{fig:vision}
\end{figure*}

    
     
\begin{figure} [ht]
	\centering
	\subfloat[Original self-attention \label{fig:oa}]{
		\includegraphics[scale=0.6]{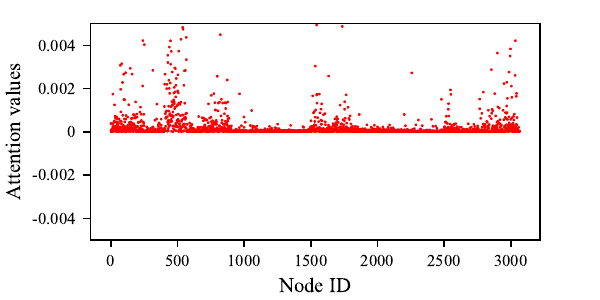}}
  \\
	\subfloat[Signed self-attention \label{fig:sa}]{
		\includegraphics[scale=0.6]{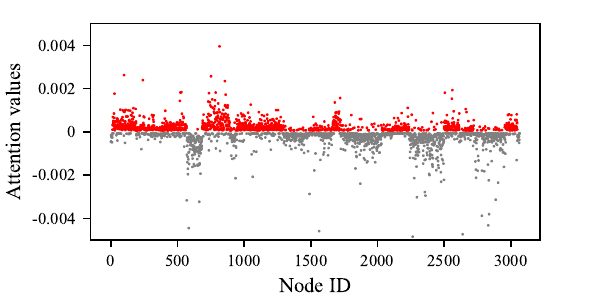}}

	\caption{The distributions of attention values learned by original self-attention mechanism and \saname.}
	\label{fig:viattention}
\end{figure}



\subsection{
Analysis on the \saname Mechanism}
To validate the effectiveness of the proposed \saname, we propose two variants of \name, named \name-O and \name-T.
The former replaces \saname with the original self-attention mechanism, while the latter utilizes $\mathit{tanh}(\cdot)$ to generate the signed attention weights.
Then, we evaluate the performance of \name, \name-O and \name-T on the node classification task. 

The results are shown in \autoref{tab:attre}.
We can observe that \name consistently outperforms \name-O and \name-T on all datasets, showing the superiority of \saname for graph representation learning.    
In addition, We can also observe that the performance of \name-T is inferior to that of \name-O on most datasets except \texttt{Actor}.
This phenomenon indicates that the signed attention mechanism derived from GNN-based models is unsuitable for Transformer-based methods. 
Moreover, we can find that the gains of \saname vary largely on different graphs.
For instance, \saname slightly increases the model performance on \texttt{Pubmed} and \texttt{UAI2010}, but significantly enhances the model performance on \texttt{Cora}, \texttt{Chameleon} and \texttt{Squirrel}. 
This observation indicates that the importance of different frequency information varies across various graphs.
And our proposed \saname can adaptively preserve different frequency information of graphs based on the semantic relevance of node pairs, which is more suitable for learning node representations on diverse graphs.

\subsection{
Analysis on  \sffn}
\sffn is another key module of \name that preserves the local topology information of $k$-hop neighborhood.
Hence, We investigate the influence of the neighborhood range $k$ on the model performance.
Specifically, we vary $k$ in $\{0, 1, 2, 3, 4, 5\}$ and observe the changes in model performance on all datasets.
Note that $k=0$ means that we remove \sffn in \name, abandoning the local topology information.

The results are shown in \autoref{fig:k}.
First, we can observe that $k=0$ leads to poor performance on all datasets, which indicates that the local topology information is also essential for learning informative node representations.
Another observation is that a large value of $k$ also impairs the model performance, which may be because that large range of neighborhoods could introduce more irrelevant information. 
Besides, although $k=1$ achieves the best performance on more than half of the datasets, several graphs, such as \texttt{Cora} and \texttt{Pubmed}, require more neighborhood information beyond immediate neighbors to achieve the best performance.
The reason is that different graphs exhibit different local topology structural features, which influence the neighborhood range differently on model performance.
This observation also indicates that flexibly capturing local neighborhood information is important for learning informative node representations on different graphs.

\subsection{Visualization}
To intuitively illustrate the advantage of \name for graph representation learning, we visualize the node representations learned by \name and the attention weights generated by the original self-attention and \saname.

\textbf{Visualization of node representations.}
We visualize the node representations extracted from the final layer of each model using t-SNE~\cite{tsne}, which is widely adopted to project the hidden representations into a two-dimensional space~\cite{bmgcn}.
Specifically, we compare \name with signed weights-based GNNs and graph Transformers on \texttt{UAI2010} as an example.
From \autoref{fig:vision} we can observe that the node representations learned by \name exhibit clearer boundaries among different labels than those learned by other methods.
It indicates that the node representations learned by \name possess higher intra-class similarity and form more discernible clusters, which is beneficial for downstream machine learning tasks.
This is because \name can adaptively learn different frequency information and simultaneously capture dependency information from global to local through two core modules, \saname and \sffn.
This further demonstrates the effectiveness of our proposed \name for learning informative node representations.

\textbf{Visualization of attention values.}
We further visualize the attention values on \texttt{UAI2010}.
Specifically, we randomly sample a node from \texttt{UAI2010} and plot the scatter diagram via attention values, as shown in \autoref{fig:viattention}.
The red circle represents the positive attention value, and the gray circle represents the negative value.
The two sub-figures depict the distributions of attention values generated by the original self-attention mechanism and our proposed \saname.
We can observe that \saname carefully preserves the influences of negative attention values.
For instance, we can find that nodes near ID 2000 exhibit very low attention values in the original attention mechanism.
In contrast, these nodes can obtain large negative attention values in \saname.
This observation also implies that \saname can capture various frequency information on graphs.

\section{Conclusion}
In this paper, we proposed \name, a novel graph Transformer for graph representation learning.
\name consists of two key modules, namely \saname and \sffn. 
\saname is a novel self-attention mechanism that generates signed attention values according to the semantic relevance of node pairs, enabling the model to adaptively preserve different frequency information between nodes.
While \sffn extends the design of FFN in Transformer by introducing the neighborhood information through the local structural bias. In this way, the local topology features could be carefully preserved.
We conducted extensive experiments on diverse benchmark datasets, including node-level and graph-level tasks.
Experimental results show that our proposed \name outperforms representative GNNs and graph Transformers on various graph mining tasks, namely node classification and graph classification, demonstrating the superiority of \name for graph representation learning on various types of graphs. 
Further ablation studies on core modules and visualizations of node representations and attention weight distributions indicate the effectiveness of our designs.

The main limitation of \name is the training cost since \name involves the attention calculation of all nodes, causing the quadratic computational complexity on the number of nodes. 
Such high training cost directly hinders the generalization of \name on large-scale graphs with millions of nodes.
Hence, optimizing the training cost of \name is one of the important future works.


\bibliographystyle{ACM-Reference-Format}
\bibliography{references}

\newpage
\appendix

\section{Datasets}\label{app_dataset}
Here, we introduce the datasets adopted for experiments according to different classification tasks.
\begin{table}[ht]
\caption{Statistics on datasets for node classification, ranked by the homophily level from high to low.}
\label{tab:ncdata}
\centering
\renewcommand\arraystretch{1.0}
\small
\scalebox{1.0}{
\begin{tabular}{lrrrrlc}
\toprule
\multicolumn{1}{l}{Dataset} & \multicolumn{1}{l}{\# nodes} & \multicolumn{1}{l}{\# edges} & \multicolumn{1}{l}{\# features} & \multicolumn{1}{l}{\# labels}  & \multicolumn{1}{c}{$\mathcal{H}$} \\ \hline
Photo& 7,650& 238,163& 745 & 8   &      0.83     \\
ACM& 3,025& 1,3128& 1,870& 3 &   0.82  \\
Cora& 2,708 & 5,278& 1,433 & 7 &       0.81      \\
Pubmed& 19,717& 44,434& 500 & 3 &      0.80   \\
Computer& 13,752& 491,722& 767& 10  &       0.78  \\
CoraFull& 19,793& 65,311& 8,710 & 70   &       0.57     \\
BlogCatalog & 5,196& 171,743& 8,189 & 3       & 0.40  \\
UAI2010& 3,067 & 28,311& 4,973& 19 &       0.36   \\
Flickr& 7,575 & 239,738 & 12,047& 9       & 0.24   \\
Chameleon& 2,277 & 31,371 & 2,325 & 5       & 0.24    \\
Actor& 7,600& 26,659& 932 & 5  &       0.22   \\
Squirrel& 5,201& 198,353& 2,089 & 5      & 0.22  \\
 \toprule
\end{tabular}
}
\end{table}

\begin{table}[ht]
\caption{Statistics on datasets for graph classification.}
\label{tab:gcdata}
\centering
\renewcommand\arraystretch{1.0}
\small
\scalebox{1.0}{
\begin{tabular}{lrrrrl}
\toprule
\multicolumn{1}{l}{Dataset} & \multicolumn{1}{l}{\# graphs} & \multicolumn{1}{l}{\# Avg. nodes} & \multicolumn{1}{l}{\# Avg. edges} & \multicolumn{1}{l}{\# labels}  \\ \hline
NCI1& 4,110& 29.87& 32.30 & 2  \\
NCI109& 4,127& 29.68& 32.13 & 2  \\
Mutag.& 4,337& 30.32& 30.77 & 2  \\
FRANK.& 4,337& 16.90& 17.88 & 2  \\
COLLAB& 5,000& 74.49& 2457.78 & 3  \\
 \toprule
\end{tabular}
}
\end{table}

\subsection{Node classification}
To evaluate the performance of \name on node classification, we adopt 12 datasets containing both homophily and heterophily graphs derived from diverse domains.
Here, we divide them into five categories: citation networks, social networks, co-purchase networks, co-occurrence networks, and Wikipedia networks.

\textbf{Citation networks.}
\texttt{Cora}, \texttt{Pubmed}, \texttt{ACM}, \texttt{UAI2010}, and \texttt{CoraFull} are citation networks, where nodes represent research papers and edges represent the citation relationships between papers.
Attribute features of nodes are bag-of-words vectors extracted from the abstract and introduction of the target paper.
Labels of nodes represent the research topics of papers.

\textbf{Social networks.}
\texttt{BlogCatalog} and \texttt{Flickr} are social networks.
Nodes represent users, and edges represent their social relationships.
Attribute features are generated based on keywords of user profiles.
And labels represent interest groups of users.

\textbf{Co-purchase networks.}
\texttt{Computer} and \texttt{Photo} are co-purchase networks extracted from the Amazon co-purchase graph, where nodes represent goods, and edges indicate that two goods are frequently bought together.
Attribute features are bag-of-words vectors generated by reviews of goods.
And labels represent the good categories.

\textbf{Co-occurrence networks.}
\texttt{Actor} is the actor co-occurrence network where nodes represent actors and edges represent actors appearing on the same Wikipedia page.
Attribute features are bag-of-words vectors generated by keywords on the Wikipedia pages.
Nodes are divided into five categories according to the words of the actor’s Wikipedia.

\textbf{Wikipedia networks.}
\texttt{Squirrel} and \texttt{Chameleon} are Wikipedia networks.
Nodes represent web pages, and edges represent mutual links between pages.
Attribute features are generated by informative nouns in the
Wikipedia pages. 
Nodes are divided into five categories based on the number of the average
monthly traffic of the web page.

The statistics of these datasets are summarized in \autoref{tab:ncdata}.

\subsection{Graph classification}
We adopt five popular datasets derived from molecules and social networks for evaluating the model performance on the graph classification task:

\textbf{Molecules.}
\texttt{NCI1}~\cite{nci}, \texttt{NCI109}~\cite{nci}, \texttt{FRANKENSTEIN}~\cite{frank} and \texttt{Muta-\\genicity}~\cite{mutag} are molecule graphs, where a graph denotes a molecule.
Nodes represent atoms, and edges represent the chemical bonds between atoms.
Labels of graphs represent toxicity or biological activity~\cite{tudataset}.

\textbf{Social networks.}
\texttt{COLLAB}~\cite{collab} is derived from scientific collaboration networks.
Graphs in \texttt{COLLAB} represent the ego-networks of researchers.
Nodes represent researchers, and edges represent the collaboration relationship of researchers.
Labels of graphs represent the research areas of researchers.

The statistics of these datasets are summarized in \autoref{tab:gcdata}.

\section{Baselines}\label{app_baselines}
We select baselines based on the following categories:

\textbf{GNNs with all positive aggregating weights.}
In this category, we choose GCN~\cite{gcn}, GAT~\cite{gat}, GraphSage~\cite{graphsage} and SGC~\cite{sgc}.
GCN is one of the classic GNN models, which leverages the Laplacian smoothing operation to obtain node representations.
GAT introduces the attention mechanism to aggregate the information information adaptively.
GraphSage leverages the node sampling strategy to reduce the training cost of GNNs.
SGC simplifies GCN by removing the non-linear activation function between GCN layers to improve the training efficiency. 

\textbf{GNNs with signed aggregating weights.}
Recently, several GNN models have utilized signed weights to aggregate the information of nodes or neighborhoods.
Here, we adopt three representative GNNs, named GPRGNN~\cite{gprgnn}, FAGCN~\cite{fagcn} and GloGNN~\cite{glognn}.
GPRGNN utilizes learnable signed weights to aggregate the information of multi-hop neighborhoods.
FAGCN modifies the GAT architecture by replacing the $\mathit{softmax}(\cdot)$ with $\mathit{tanh}(\cdot)$ to enable the attention mechanism to generate signed attention values.
GloGNN leverages the learnable coefficient matrix with positive and negative values to aggregate the information of nodes.

\textbf{Graph pooling-based methods.}
Graph pooling is a popular technique to learn the representations of graphs for graph-level data mining tasks.
Here, we select three representative graph pooling-based methods as baselines, Set2Set~\cite{set2set}, SAGpool~\cite{sagpool}, and SortPool~\cite{sortpool}.

\textbf{Graph Transformers.}
We also adopt representative graph Transformers as baselines.
Specifically, we adopt GT~\cite{gt}, GraphTrans~\cite{graphtrans}, Graphormer~\cite{graphormer}, SAT~\cite{sat}, NodeFormer~\cite{nodeformer} and Specformer~\cite{specformer}.
GT utilizes the adjacency matrix to sparse the attention matrix.
GraphTrans and SAT leverage GNN-based methods to extract the structural information of nodes.
Graphormer develops three encoding strategies, including centrality encoding, spatial encoding and edge encoding, to capture the structural information of graphs.
NodeFormer and Specformer are recent advanced Transformer-based models for node classification.
NodeFormer introduces the linear self-attention mechanism to reduce the training cost.
Specformer leverages the self-attention mechanism to encode the relations of eigenvalues in the spectral domain. 
We also adopt the standard Transformer~\cite{transformer} as the baseline.

\section{Implementation details}\label{app_para}
Here, we introduce the implementation details of experiments, including experimental environments and parameter settings.

\textbf{Experimental environment.}
All experiments are conducted on a Linux server with one Xeon Silver 4210 CPU and four RTX TITAN GPUs. 
Codes are implemented on Python 3.8 and Pytorch 1.11.

\textbf{Parameter settings.}
For each baseline, we adopt the official implementation on GitHub.
Referring to their recommended settings, we perform hyper-parameter tuning for all models.
For the proposed \name, we try the number of layers in $\{1,2,3\}$, the number of hidden dimension in $\{128, 256, 512\}$, the range of $k$-hop neighborhood in $\{1,2,3,4,5\}$.
Model parameters are optimized by AdamW \cite{adamw} with the learning rate in $\{1e-2,5e-3,1e-3\}$, the weight decay in $\{1e-4,5e-4,1e-5\}$ and the dropout rate in $\{0.1,0.3,0.5\}$.




\end{document}